%% file: main.tex
\documentclass[10pt,letterpaper]{article}

\usepackage{cogsci}

\cogscifinalcopy % Uncomment this line for the final submission 

\usepackage{pslatex}
\usepackage{apacite}
\usepackage{float} % Roger Levy added this and changed figure/table
\usepackage{setspace}

\usepackage{latexsym}
\usepackage{url}
\usepackage[T1]{fontenc}

\usepackage{amsmath}
\usepackage{comment}
\usepackage{graphicx}
\usepackage{float}
\usepackage[algo2e]{algorithm2e}
\usepackage{soul}

\usepackage{tabularx}

\usepackage[normalem]{ulem}
\usepackage{multirow}
\usepackage{caption}
\usepackage{subcaption}

\usepackage{xspace,mfirstuc,tabulary}

\usepackage{cleveref}
\usepackage{lineno}
\usepackage{inconsolata}
\usepackage{makecell}

\usepackage{gb4e}
\noautomath

\usepackage{xcolor}

\newcommand{\EXTENSION}[1]{}
\newcommand{\frameworkx}{NeLLCom-X}
\newcommand{\framework}{NeLLCom}

\newcommand{\typea}{\neg Amb}
\newcommand{\typeb}{Amb}

\title{Simulating the Emergence of Differential Case Marking \\ with Communicating Neural-Network Agents}
 
\author{
  Yuchen Lian$^\diamond$ $^\dagger$
  \qquad
  Arianna Bisazza$^\ddagger$
  $^{\ast}$
  \qquad
  Tessa Verhoef$^\dagger$\thanks{Shared senior authorship.}  \\
  \ \\
  $^\diamond$Faculty of Electronic and Information Engineering, Xi'an Jiaotong University
  \\
  $^\dagger$Leiden Institute of Advanced Computer Science, Leiden University
  \\
  \texttt{\{y.lian, t.verhoef\}@liacs.leidenuniv.nl}
  \\
  $^\ddagger$Center for Language and Cognition, University of Groningen \\
  \texttt{a.bisazza@rug.nl}
}

\begin{document}

\maketitle

% \makeatother
% \renewcommand{\thefootnote}{\arabic{footnote}}

\begin{abstract}
%\AB{I think we should spend some words on DCM and what makes it interesting among many other phenomena we could have chosen}
%\TV{I wrote a new abstract and I usually try to avoid putting citations in the abstract, but I am now thinking that this leads to a description of the previous human experiments that is a bit too generic, so perhaps we should at least cite S\&C in the abstract?}

Differential Case Marking (DCM) refers to the selective use of grammatical case marking based on semantic, pragmatic, or other factors. The emergence of DCM has been studied in artificial language learning experiments with human participants, which were specifically aimed at disentangling the effects of learning from those of communication \cite{smith2020communicative}. Meanwhile, multi-agent reinforcement learning frameworks based on neural networks have gained significant interest to simulate the emergence of human-like linguistic phenomena.
In this study, we employ such a framework in which agents first acquire an artificial language before engaging in communicative interactions, enabling direct comparisons to human results. Using a very generic communication optimization algorithm and neural-network learners that have no prior experience with language or semantic preferences, our results demonstrate that learning alone does not lead to DCM, but when agents communicate, differential use of markers arises. This supports \citeA{smith2020communicative}'s findings highlighting the critical role of communication in shaping DCM and showcases the potential of neural-agent models to complement experimental research on language evolution.

%Recent advancements in computational linguistics have leveraged neural networks to model human communication. 
%Multi-agent reinforcement learning frameworks based on neural networks have gained significant interest to simulate and study the emergence of human-like communication protocols \AB{(cite Boldt?)}.
%In the recently introduced NeLLCom framework \cite{lian-etal-2023-communication,lian-etal-2024-nellcom}, agents first learn a pre-defined artificial language and then use it for communication, enabling direct comparisons to human results from Artificial Language Learning (ALL) experiments.
%
%We adopt this framework to investigate the Differential Case Marking (DCM) phenomenon, closely replicating the design of previous human ALL experiments \AB{which were specifically aimed at disentangling the effects of learning from those of communication}. 
%The results of our simulations successfully replicate the DCM effect %after communication 
%\AB{using a very generic communication optimization algorithm} ... \AB{+  
% Connect this to S\&C findings about comm vs learning?}
%even with modifications toward a more neutral language setting. 
%\AB{If we have a good sentence here, we can remove the next one}
%
%This work serves as a strong supplementary example of the accessibility and controllability of combining Artificial Language Learning and the NeLLCom framework to study linguistic phenomena.

\end{abstract}

\section{Introduction}

Human language is not a static entity but a dynamic system undergoing continuous change and evolution. %Its linguistic structure is shaped by mechanisms operating across different time-scales. On shorter time scales, interaction and communication facilitate the negotiation of new meanings, while on longer time scales, processes such as learning and transmission across generations give rise to emergent linguistic patterns and enhance learnability. Many distinctive features of human language can be understood as adaptations to the contexts in which language is used and transmitted.
The development of agent-based models is a productive approach to studying the emergence and change of linguistic systems, which has a long-standing tradition in the study of language evolution \cite{boer2006computer, steels1997synthetic, hare1995learning, hurford1989biological}. Recent advancements in computational linguistics and deep learning have reinvigorated interest in such simulations, providing the opportunity to model increasingly realistic phenomena
\cite{chaabouni2021communicating, lian-etal-2021-effect, lian-etal-2023-communication}. These models simulate the spontaneous development of communication systems through repeated interactions among individual neural-network agents 
% \cite{lazaridou2017multiagent, havrylov2017emergence, chaabouni-etal-2020-compositionality}
(\citeA{lazaridou2017multiagent, havrylov2017emergence, chaabouni-etal-2020-compositionality, boldt2024review}. %\lyc{see \citeA{boldt2024review} for a comprehensive survey})
A key challenge in this area is that the languages developed by these agents, when initialized from scratch, often lack human-like characteristics \cite{chaabouni-2019-antiefficient, galke2022emergent, lian-etal-2021-effect}. 
%
%%% AB: Removed as less relevant in the context of studying HUMAN language:
%Moreover, when agents are initially trained on natural human languages, they may drastically diverge from human-interpretable utterances after several interaction rounds, leading to language drift \cite{lee-etal-2019-countering,lazaridou2020multi, Lowe2020On}. 
%%%
An alternative approach is employed in the Neural agent Language Learning and Communication (NeLLCom) framework \cite{lian-etal-2023-communication}, where agents first learn a predefined miniature artificial language and then use it for communication, with the goal of studying the emergence of specific linguistic properties.

For instance, NeLLCom has been used to investigate the emergence of a trade-off between case-marking and word order strategies \cite{lian-etal-2023-communication, lian-etal-2024-nellcom}, a phenomenon commonly observed in natural languages. Case marking and word order are both strategies to indicate who does what to whom in a sentence and languages often rely more heavily on one strategy than the other. This trade-off had previously been found to emerge in artificial language learning (ALL) experiments with humans \cite{fedzechkina2017balancing}, where participants dropped the use of markers more often if they were learning artificial languages with fixed word order than in the case of flexible word order. %Such an effect did not emerge, however, in initial simulations with neural network agents \cite{chaabouni-etal-2019-word, lian-etal-2021-effect}.
Having adapted the experimental design and artificial languages of \citeA{fedzechkina2017balancing} to train neural network agents, \citeA{lian-etal-2023-communication} found human-like patterns of language change only
when agents actively attempted to be understood by a communication partner. Communication provided a pressure for the language to develop towards a form that makes it maximally efficient without losing communicative success. Simplifying the language by dropping the markers was possible when word order provided enough cues to derive the meaning correctly. 

In naturally occurring human languages, word order is not the only factor that may influence the efficient use of markers. Even in languages with flexible word order, case-marking is frequently employed selectively rather than universally \cite{witzlack2018differential, sinnemaki2014typological, de2008case, levshina2021communicative}. This paper focuses on Differential Case Marking (DCM), a widespread phenomenon observed across many languages, where the morphological marking of a grammatical case varies depending on semantic, pragmatic or other factors. For example, in many languages, animate objects are explicitly marked to clarify their role in a sentence \cite{garcia2018nominal, levshina2021communicative} since animate entities typically appear in the subject instead of the object role (e.g. in an event involving \textit{eating}, \textit{cake} and \textit{Alice}, we typically assume Alice to be doing the eating). The emergence of this phenomenon has been explored through various ALL experiments with human participants (\citeA{smith2020communicative}, henceforth S\&C; \citeA{fedzechkina2012language}, henceforth FJN), particularly in relation to the roles of learning and communication which S\&C explicitly sought to disentangle.
% \cite{smith2020communicative, fedzechkina2012language}. 
This makes the DCM phenomenon highly suitable for investigation using the NeLLCom framework, %\cite{lian-etal-2023-communication, lian-etal-2024-nellcom}, 
where effects of communication and learning can be simulated with a generic communication protocol and linguistically naïve learners, while closely replicating the experimental setups and language design of FJN and S\&C. Our results provide additional evidence supporting S\&C's proposal that communication plays a crucial role in shaping the emergence of DCM.

%In this study, we replicate their experimental setups using NeLLCom \cite{lian-etal-2023-communication, lian-etal-2024-nellcom} to provide additional evidence supporting 
% Smith and Culbertson's \citeyear{smith2020communicative}
%S\&C's proposal that communication plays a crucial role in shaping the emergence of DCM.
%\AB{we could slightly rephrase this to emphasize even more that this focus on comm vs learning in ALL human experiments makes it an especially good fit for the NeLLCom framework}

%Although the word order/case-marking trade-off is often regarded as a linguistic universal,  The emergence of this phenomenon has been explored through ALL experiments with human participants \cite{smith2020communicative, fedzechkina2012language}, and in this study, we replicate their experimental setups using NeLLCom-X to provide additional evidence supporting Smith and Culbertson's proposal that communication plays a crucial role in shaping the emergence of DCM.

\section{Differential Case Marking}
\label{src:dcm}

Differential case marking, or differential argument marking, refers to a widespread cross-linguistic phenomenon in which the formal marking strategy for an argument differs according to its semantic, pragmatic, or other properties \cite{witzlack2018differential, sinnemaki2014typological, de2008case, levshina2021communicative}. More specific instances of DCM are Differential Object Marking (DOM) and Differential Subject Marking (DSM). An example (adapted from \citeA{garcia2018nominal, levshina2021communicative}) of DOM is the marking of human objects in Spanish:
\begin{exe}
\ex
\begin{xlist}
\item \gll \textit{Pepe} \textit{ve} \textit{la} \textit{película.} \\
     Pepe sees the film. \\
\glt ‘Pepe sees the film.’
\item \gll \textit{Pepe} \textit{ve} \textit{a} \textit{la} \textit{actriz.} \\
     Pepe sees TO the actress. \\
\glt ‘Pepe sees the actress.’
\end{xlist}
\end{exe}
where the atypical animate/human object in (b) is marked using \textit{`a'}. Similarly, in DSM, typical subjects can remain unmarked while atypical subjects are more likely to be marked.

There is a long-standing debate about the mechanisms that cause this phenomenon to develop. 
Typological studies \cite{aissen2003differential,croft2003typology,levshina2021communicative}, artificial language experiments \cite{fedzechkina2012language,smith2020communicative,tal2022impact}, and computational simulations \cite{lestrade2018emergence} have been conducted to explore potential explanations. \citeA{levshina2021communicative} broadly contrasts two explanations for the emergence of DCM, which have been previously discussed in the literature. The first considers this phenomenon to be a result of efficient communication strategies, where markers are used more in cases where the probability to be misunderstood without them is higher. The second invokes markedness theory, where unmarked linguistic forms (the default or neutral forms that are more frequent and simpler in a given context) tend to be used for typical events, while (more atypical and complex) marked forms are iconically associated with atypical events. Corpus-based quantitative analyses suggest that the first (efficient communication) is a better predictor of cross-linguistic patterns observed in DCM \cite{levshina2021communicative}.
%\AB{REMOVE THIS PART after moving a short version to the intro:}
%In the context of agent-based modeling,
%\citeA{lestrade2018emergence} simulated the emergence of DCM %by designing a model named \textit{WDWTW} (for \textit{who does what to whom}) in which 
%with relatively simple agents in a model where %communicate with each other using words from an initial lexicon, modeled as a list of randomly generated vectors. M
%marking strategies, heuristics for interpreting messages and grammaticalization principles were explicitly built-in to see what their combined or separate impact was. % on the emergence of DCM.  
%Generations of agents iteratively operate within a virtual environment to achieve successful communication. 
%Each agent maintains an evolving lexicon, common ground, and usage history.
%The lexicon is modeled as a list of randomly generated 
%vectors. The dimensions can be interpreted as any grammatically relevant properties of words in natural language, but the model does not commit to any specific interpretation of them.
%Conversation is guided by a set of shared consistent communication principles.
\citeA{lestrade2018emergence} simulated the emergence of DCM with relatively simple interacting agents in a model where (in contrast to our work) marking strategies %, heuristics for interpreting messages
and grammaticalization principles were explicitly built-in to see what their combined or separate impact was. Their results suggested that argument marking can evolve gradually as languages adapt to usage. %, where 
%Rooted in communicative need, a differential marking system emerged before a full case-marking system arises where marking is obligatory (this last stage was not reached in the simulation though). %Our neural agent simulations differ from this work in that we do not implement marking strategies or principles of grammatical change explicitly, but rather create a domain-general environment where repeated communication can result in linguistic change. % and the emergence of different strategies happens spontaneously.
%Note that, unlike neural agent simulations, agents in \textit{WDWTW} produce and process utterances using the same shared basic protoprinciples, such as Typing, Grouping, AgentFirst, and CheckSuccess, even though their lexicons are distinct. In contrast, neural agents not only differ in their word and meaning representations but also possess unique speaking and listening strategies, determined by their individual network parameters.

%, such as efficiency \footnote{explain this account.} and markedness \footnote{explain this account.}.
% Among various possible accounts, the most prominent accounts involved by these studies are efficiency and markedness \cite{levshina2021communicative}.
%
%Focusing on animacy, the most relevant features in DAM, 
In a laboratory setting, FJN conducted a set of ALL experiments, where human participants watched computer-generated videos and heard their descriptions in a novel artificial language.
%A set of two inefficient verb-final languages with flexible word order (SOV dominant with 60\% of all sentences) and optional case marking (overall 60\% case marked) is provided as initial input languages. 
%Here, case marking is independent of the animacy of the argument, which deviates from natural languages.
After 4 days of learning, researchers found that the learners' productions deviated from their input language towards more efficient case-marking systems (using markers more often for atypical arguments like animate objects or inanimate subjects, than in typical situations).
%Specifically, for a language with object marking, they produced significantly more case markers on atypical (animate) objects than on typical (inanimate) objects across all days of testing, mirroring the common DOM pattern.
%For a subject marking language, learners used more case marking on atypical (inanimate) subjects than on typical (animate) subjects on the final day of training, matching the DSM pattern. 
The authors therefore conclude that language learners restructure their linguistic input so that it increasingly facilitates efficient communication. 
As pointed out by S\&C, this interpretation is surprising, since it is more typically assumed that language \textit{use}, and not \textit{learning}, drives its evolution towards communicative efficiency \cite{kirby2015compression, kemp2018semantic, gibson2019efficiency}. 
%Besides this conceptual issue about the relative role of learning and communication, the event configuration is also debatable for encouraging the DCM-like effect according to S\&C. 
Perhaps even more surprising, the language design by FJN was such that in the 10 animate nouns in the lexicon of the object marking language for example, 5 only occur as subjects and the other 5 as objects.
The meaning was therefore potentially unambiguous regardless of the presence of a marker, which conflicts with the efficiency account (where disambiguation for a listener is assumed to drive the effect). Notably, these results were also consistent with an explanation based on markedness theory and iconicity instead of efficient communication (S\&C).
% \cite{smith2020communicative}.
S\&C adapted and extended the experiments of FJN, in a large-scale study (n > 300), and introduced an interaction phase after the last day of learning where participants use the language to communicate with a simulated interlocutor implemented as a simple chatbot. Their findings do not replicate those of FJN. Instead, they suggest that learning alone cannot reliably explain the emergence of DOM, but actual communicative interaction is key to the emergence of a communicatively-efficient case marking system. Complementing these findings, we simulate neural-agent learning and communication with agents that do not have any iconic preference, linguistic knowledge or sense of animacy. 
%to investigate whether a DCM effect emerges in these two scenarios when arguments appear in (statistically) typical or atypical roles.
This allows us to investigate whether typicality alone can lead to a DCM effect, and whether communicative pressures are a necessary factor for the emergence of DCM in neural learners, like in humans.

\section{\framework\ Framework}

Artificial language learning has been widely used in experiments with humans \cite{culbertson_all_2023, fedzechkina2016miniature}, as it provides a means to isolate specific linguistic phenomena and study cause-and-effect relationships in a controlled setting. These human-based studies can serve as valuable inspiration for the design of emergent communication simulations, allowing for direct comparisons between human and agent behaviors. 
The \framework\ framework \cite{lian-etal-2023-communication,lian-etal-2024-nellcom} is a multi-agent communication framework designed to simulate ALL experiments for the study of language change and evolution. %through pre-defined artificial languages. 
After being trained on an initial artificial language through supervised learning, agents in this framework start interacting via meaning reconstruction games in which they optimize a shared communicative reward through reinforcement learning. 
%While ALL has also been applied in the context of neural network learners—particularly to explore the inductive biases of neural learners—such studies often focus solely on passive learning. In contrast, NeLLCom emphasizes the interactive nature of language by following the learning phase with a Communication Phase, as is common in many emergent communication models.

NeLLCom enables researchers to scale ALL experiments in ways that are difficult to achieve with human participants. While \citeA{fedzechkina2017balancing} focused solely on individual learning by human participants, \citeA{lian-etal-2024-nellcom} recently expanded their work on the word order/case-marking trade-off using NeLLCom-X, incorporating more realistic role-alternating agents and group communication. This extension demonstrated that the trade-off also emerges in populations of communicating individuals, which is something that would be rather difficult and expensive to achieve with human participants in a lab.
Here we use the most recent version of the framework, NeLLCom-X.

\paragraph{The Task}
In \frameworkx\, \textbf{meanings} describe simple scenes using triplets $m=\{Action,\ agent,\ patient\}$
% , where $A$, $a$, $p$ correspond to an action, an agent, and a patient, respectively
(e.g., \textsc{eat, alice, cake}). 
An artificial \textbf{language} is defined by a set of grammatical rules generating utterances $u$ from a fixed-size vocabulary to convey meaning $m$. 
Utterances can be of variable length and multiple $u$ candidates can be valid for the same $m$.
In the meaning reconstruction game, a speaker conveys a meaning $m$ by generating an utterance $\hat{u}$, which the listener then maps to meaning $\hat{m}$. The game is successful if $m=\hat{m}$.

\subsubsection{Agent Architecture}

The structures of listening and speaking networks are symmetric with meanings represented by unordered tuples while utterances are generated/processed sequentially. This results in a \textbf{linear-to-RNN} (Recurrent Neural Network) speaking network \mbox{$\mathcal{S}: m \mapsto u$} and a \textbf{RNN-to-linear} listening network \mbox{$\mathcal{L}: u \mapsto m$}. An agent then includes two sets of parameters \mbox{$\alpha_{i}=(N_{i}^\mathcal{S},N_{i}^\mathcal{L})$} tied together through parameter sharing of their meaning and word embeddings.

% The \textbf{speaking} network, with a linear-to-RNN structure, implements the function \mbox{$\mathcal{S}: m \mapsto u$}, while the \textbf{listening} network, with a symmetric RNN-to-linear structure, implements the function \mbox{$\mathcal{L}: u \mapsto m$}.
% In both directions, meanings $m$ are represented by unordered tuples instead of sequences to avoid ordering bias.

%Through \textbf{parameter sharing}, \frameworkx\ allows an agent to take both roles (speaker and listener), which is a key ingredient for realistic simulations of emergent communication \cite{galke2022emergent}. 
%Specifically, this symmetric structure of the listening and speaking networks allows tying the meaning embeddings and the word embedding respectively. 
% \begin{align}
%     \mathbf{X}(N_i^\mathcal{S}) = \mathbf{O}(N_i^\mathcal{L}) \\
%     \mathbf{X}(N_i^\mathcal{L}) = \mathbf{O}(N_i^\mathcal{S})
% \end{align}

\subsubsection{Training}
Before communication, 
each agent is first trained by Supervised Learning (\textbf{SL}). Using a set of reference meaning-utterance pairs $D={(m,u)}$ and teacher forcing, this phase minimizes the cross-entropy loss between $u$ and the words generated by the speaker given $m$. 
Conversely, for the listener, SL minimizes the loss between $m$ and the meaning tuple generated by the listener given $u$.
Then, two 
(or more\footnote{We only consider two-agent communication in this work. However \frameworkx\ can model group communication with more than two individuals by iteratively sampling pairs of two agents from the group to proceed with an interaction.})
trained agents $\alpha_0$ and $\alpha_1$ learn to communicate with each other via Reinforcement Learning (\textbf{RL}). During this phase, agents maximize a shared communication reward $r(m,\hat{u})$ which captures how close the listener’s prediction  $\mathcal{L}(\hat{u})$ given the speaker-generated utterance $\hat{u}=\mathcal{S}(m)$ is to $m$. %$\mathsf{comm}(\alpha_0,\alpha_1) =\mathsf{RL}(N^\mathcal{S}_0,N^\mathcal{L}_1)$.
See more details in \cite{lian-etal-2024-nellcom}.\footnote{
In each interaction turn, each agent is assigned to a role (speaker or listener) with equal probability. %Here $\alpha_0$ is a speaker and $\alpha_1$ is a listener.
To ensure self-understanding is maintained, rounds of %$\mathsf{comm}$ 
interaction between different agents are interleaved at regular intervals with rounds of \textit{self-communication} where an agent's speaking network sends messages to its own listening network.}

%  SHORTENED AND MOVED TO FOOTNOTE ABOVE:
% Besides the $\mathsf{inter\_comm}$ which is between a speaking agent and a different listener agent, $\mathsf{self\_comm}$ is implemented by letting the agent's speaking network send messages to its own listening network while optimizing the shared reward. $$\mathsf{self\_comm}(\alpha_i) =\mathsf{RL}(N^\mathcal{S}_i,N^\mathcal{L}_i)$$
%This $\mathsf{self\_comm}$ (also known as self-play in \citeA{lowe2019interaction,lazaridou-etal-2020-multi}) is a procedure necessary to preserve the agents' self-understanding while their language evolves in interaction.
%\footnote{Even when word and meaning representations are shared, the rest of the speaking and listening networks remain disjoint, potentially causing the speaking and listening abilities to drift in different directions.}

%%%%%%%%%%%%%%%%%%%%%%%%%%%%%%%%%%%%%%%%%%%%%%%%%%%%%%%%%%%%%%%%%%%%%

\section{Experimental Setup}
We use \frameworkx\ to simulate the emergence of DCM in neural agents, following the language design of FJN and S\&C as explained in this section.

\subsubsection{Meaning Space}
As previously mentioned, %in Sect. \ref{src:dcm}, 
DCM implies that marker production can differ depending on the typicality of the entities in a sentence. 
Mirroring human languages where animate agents (e.g. \textit{Alice}) and inanimate patients (e.g. \textit{cake}) are more typical, 
the \textbf{Object-Marking} condition in FJN and S\&C %in human lab experiments \cite{fedzechkina2012language,smith2020communicative} 
defines a meaning space where
agents are always animate entities, while patients can be either animate or inanimate.
% SEEMS OBVIOUS TO ME: (an animate entity can be either an agent or a patient, while an inanimate entity only occurs in the patient role).
Conversely, in the \textbf{Subject-Marking} condition, patients are always inanimate, while agents can be either animate or inanimate. % and inanimate patients can also appear as agents. 
%(an inanimate entity can be either an agent or a patient, while an animate entity only serves as an agent).
% NO NEED TO MENTION THIS, AS IT'S MENTIONED BELOW IN THIS SECTION: \AB{Following the adjusted design of S\&C, the same entity ...}
Note that, in the human experiments, animacy referred to a property of the entities depicted in the stimuli, which were concepts previously known to the participants (e.g. animate \textit{artist, baker}, etc. versus inanimate \textit{ball, cake}, etc.).
By contrast, neural networks are trained from scratch and have no previous world knowledge.
%All entities are therefore encoded the same way in neural networks, and 'animacy' is only inferred from the statistical properties of the language. 
%Instead of complicating the meaning encoding system with an additional animacy feature, we simulate the typicality/ambiguity difference between entities by different entity distributions.
In our setup, all entities are encoded in the same way (as entries of the meaning embedding table, all randomly initialized), and the typicality of an entity's role is inferred from the statistical properties of the observed meaning space (e.g. `entity-3' occurring half of the times as agent and the other half as patient, versus `entity-5' occurring always as patient).
Working with neural agents therefore allows us to tease apart the effect of typicality as a purely statistical property from prior animacy associations, which was not possible in the setup of S\&C's or FJN's human experiments. % (where humans were exposed to known entities)
%In our meaning triplets $m=\{Action,\ agent,\ patient\}$, $agent$ and $patient$ are entities with typicality features.
We call $\typeb$ (ambiguous) the subset of entities that can have two roles, and $\typea$ (unambiguous) the subset of entities that can only occur in one role.
Thus, possible meaning structures are $\{A, a_{\typeb}, p_{\typea}\}$ and
$\{A, a_{\typeb}, p_{\typeb}\}$
in the Object-Marking language;
$\{A, a_{\typea}, p_{\typeb}\}$ and
$\{A, a_{\typeb}, p_{\typeb}\}$
in the Subject-Marking language.%
\footnote{Note this setup corresponds to the `Subjects Can Be Objects' condition introduced by S\&C.}

In each language condition, 20 entities (10 ambiguous and 10 unambiguous) and 8 actions are included, 
resulting in a total of $10 * (10 + (10-1)) * 8 = 1520$ possible meanings.
This expanded meaning space results in a better model convergence in preliminary experiments \cite{zhao2018bias,chaabouni-etal-2020-compositionality}, compared to the relatively small space used in human experiments (10 entities and 4 verbs).
%\footnote{$a$ and $p$ cannot be the same entity in a meaning.}

%\subsubsection{Word Order and Case Marking}
\subsubsection{Artificial Languages} 

\renewcommand{\arraystretch}{1}
\begin{table}[h]
\input{contents/lang_table_v4}
\end{table}

\newcommand{\subfigsize}{.242}
\begingroup
\setlength{\tabcolsep}{0pt} % Default value: 6pt
\renewcommand{\arraystretch}{0.9}
\begin{table*}[h!]
\input{contents/fig_merge_v2.tex}

\captionof{figure}{%\TV{I still didn't like the way the lines were drawn in this table, the figures were partly drawn over the lines, making them thicker at the start, end and between the figures and the whole table looked very 'loud', so for now I removed the horizontal lines and cropped out the x-axis label of the first and second row of images. 
% \TV{I think it is not perfect yet, but at least cleaner than before?}\AB{looks good to me}
Agent communication results for the three initial languages. % shown in Table~\ref{tab:language-table}. 
\textbf{Column~1}: Meaning reconstruction accuracy across communication rounds, computed on the whole test set (orange line), as well as split by non-ambiguous (green) and non-ambiguous (blue) meanings. 
\textbf{Col.~2-4}: Production preferences (PP) in terms of order proportions and marker use.
Specifically, Col.~2 and 3 show PP for non-ambiguous and ambiguous meanings respectively, before and after communication, and Col.~4 shows the difference in PP between $M_{test,\ \neg Amb}$ and $M_{test,\ Amb}$ after communication.
Solid diamonds mark the initial
language. Each empty circle is an agent and solid circles are the average of all agents, with error bars showing standard deviation.
%Pinks represent the before-communication production, purples represent the after-communication production. I THINK IT'S ENOUGH TO SAY THIS IN THE LEGEND
Each experiment is repeated with 50 agent pairs. %\AB{(i.e. 100 random seeds)}.
}\label{fig:comm_result}
\caption*{}
\end{table*}
\endgroup

Following FJN and S\&C, we adopt verb-final languages allowing SOV or OSV orders in varying proportions.
The token \textit{`mk'} serves as a case marker and is optionally assigned to either the subject or object based on the language type. 
%Note that the predefined marker and order proportions are not conditioned on the entity animacy.
For example, given the meaning $m$=\{$A$: \textsc{eat}, $a$: \textsc{alice}, $p$: \textsc{cake}\}, flexible-order object-marking languages admit four utterances: 
\textit{`Alice cake eat'}, 
\textit{`Alice cake mk eat'},
\textit{`cake Alice eat'}, and
\textit{`cake mk Alice eat'}.
 
A specific language is defined by four factors: 
whether it is object- or subject-marking, 
its order profile $p(SOV)$, 
marking proportion in the SOV order $p(mk|SOV)$, and 
marking proportion in the OSV order $p(mk|OSV)$.
%
%Thus 4 factors can define a specific language: $l = \{type,\ p_{SOV},\ (p_{mk|SOV},\ p_{mk|OSV})\}$. 
%The general marker proportion $p_{mk} = p_{SOV} * p_{mk|SOV}+ (1 - p_{SOV}) * p_{mk|OSV}$.
We consider two types of languages. 
The first \textbf{dominant order language}
%$l_{dom-OBJ} = \{Obj,\ 60\%\ p_{SOV},\ (67\%\ p_{mk|SOV},\ 50\%\ p_{mk|OSV})\}$ 
replicates one of S\&C's target languages, which was in turn designed following FJN. %is the replication of Experiment 1 in S\&C. 
This is an object-marking language 
with $p(SOV)=60\%$, 
$p(mk|SOV)=67\%$, and
$p(mk|OSV)=50\%$,
%with 67\% percentage of objects being marked, the minor 40\% OSV order with half of the objects being marked,
resulting in an overall 60\% marking proportion. This language was designed by FJN to simulate real flexible-order languages, where one order is typically dominant.
However, a limitation of this design is that neural agents may amplify the initial bias towards using more SOV order and marking SOV utterances more often than OSV ones, driven by a generic pressure to regularize their input. 
We thus expect agents to drift towards a strongly SOV and strongly SOV-marking solution, just because those were the most frequently observed patterns in the training data.
%result on this language is insufficient to represent the inherent bias of the neural agents.

%\AB{==DONE TIL HERE}

To disentangle input bias amplification from the actual agents' preferences towards different DCM strategies, we also experiment with a \textbf{neutral order language}, where SOV and OSV are evenly distributed, and marking proportion is 67\%.\footnote{In preliminary experiments starting from 50\% case marking, we observed a strong tendency of the agents to drop markers altogether, making it impossible to explore the DCM effect.}
We implement both an object-marking and a subject-marking version of this language. 
Table~\ref{tab:language-table} summarizes the three languages used in our experiments.
%s with neutral distributions of word order, namely 50\% SOV and 50\%OSV, and 67\% case marking proportion for each order.

% \begin{align}
%     l_{ntl-Obj} = \{Obj,\ 50\%\ p_{SOV},\ (67\%\ p_{mk|SOV},\ 67\%\ p_{mk|OSV})\} \\
%     l_{ntl-Subj} = \{Subj,\ 50\%\ p_{SOV},\ (67\%\ p_{mk|SOV},\ 67\%\ p_{mk|OSV})\}
% \end{align}

% The two languages only differ in the marking constituent, namely Object-marking language ($l = \{Obj,\ p_{SOV}=50\%,\ (p_{mk|SOV}=67\%,\ p_{mk|OSV}=67\%)\}$) and Subject-marking language ($l = \{Subj,\ p_{SOV}=50\%,\ (p_{mk|SOV}=67\%,\ p_{mk|OSV}=67\%)\}$).

Following \citeA{lian-etal-2024-nellcom}, each entity corresponds to a word, resulting in the fixed-size vocabulary = 20 + 8 + 1 = 29.

\subsection{Evaluation}

Accuracy and production preference evaluation are adopted from 
\citeA{lian-etal-2023-communication,lian-etal-2024-nellcom}.
All evaluations are based on an unseen meaning set $M_{test}$.
In the SL phase, performance is measured by listening and speaking accuracy against the reference dataset. 
In the RL phase, communication success %, which includes self-understanding and interactive communication success, 
is evaluated by meaning reconstruction accuracy, where $acc(m, \hat{m})$ equals 1 iff the entire meaning is matched.
Production preferences are visualized as the proportion of markers and different orders in a set of utterances generated by an agent for $M_{test}$, after discarding utterances that are not well-formed according to the language grammar.\footnote{In our experiments, the ratio of non well-formed utterances averaged over seeds is around 10\% before RL and 21-25\% (depending on the initial language) after RL, which is overall comparable to the results in \cite{lian-etal-2024-nellcom}.}
%\footnote{\AB{Following \citeA{lian-etal-2024-nellcom}, only utterances that are well-formed according to the grammar are taken into account.}}.
Furthermore, we split $M_{test}$ into ambiguous 
$M_{test,\ Amb}$ and unambiguous $M_{test,\ \neg Amb}$ meanings, 
%$M_{unambiguous}$ and unambiguous $M_{ambiguous}$ meanings, 
% based on the difference of ambiguity, 
and evaluate the two sets separately. 

We use generalized linear mixed-effects models (GLMMs) in the lme4 package version 1.1-35 \cite{lme4} with R Version 4.4.2 \cite{r442} to evaluate how the marking proportion and word order preferences are influenced by ambiguity after communication and whether marking use is conditioned on word order.
% \lyc{In addition, we adopt the simple linear regression model (lm from R) to evaluate whether an existing effect is linear.}

\subsection{Model Configuration and Training Details}
% we have real communication instead of a simulator

We apply the same configuration as \citeA{lian-etal-2024-nellcom}. The sequential layer in speaking and listening networks consists of 16-dimensional Gated Recurrent Units \cite{chung2014empirical}. The shared meaning embeddings and shared word embeddings have 8 and 16 dimensions, respectively.
Utterance length for the speaker is limited to 10 words.

We first split the dataset $D$ into 80/20\% training/test samples. 
The test split (304 meanings) is used throughout the whole evaluation. 
During SL,
we resample 66.7\% meanings out of the first train set following the all-seen-entities/actions rule \cite{lian-etal-2023-communication}.
SL iterates 60 times with a 0.01 learning rate using the default Adam optimizer.
During RL, 
320 meanings are sampled from the first train set and used as the training samples for each communication turn.
RL iterates 200 inter\_turns
% (resulting in 20 self\_turn for each agent) 
with a 0.005 learning rate using the \textsc{reinforce} algorithm \cite{williams1992simple}.
Batch size is set to 32 in both SL and RL training. 
We repeat each language setup with 50 pairs of agents (i.e. 100 random seeds). 

%\section{Human Experiment Replication}
\section{Results}
%All results are presented in Figure~\ref{fig:comm_result} and discussed below.

\subsection{Dominant Order Language}
%We first implement the dominant order language  used in S\&C.
% :$L_{Obj}[60_{SOV}\&(67_{Smk}+50_{Omk})]$, 
% where the SOV takes the majority order of 60\%, with 67\% percentage of objects being marked, 
% the minor 40\% OSV order with half of the objects being marked, resulting in an overall 60\% marking proportion.
% \footnote{This biased language is designed to simulate real human languages according to FJN.}

%\subsubsection{After SL}

Results for the language adopted from S\&C are presented in the \textbf{first row} of Figure~\ref{fig:comm_result}.
\textbf{Before the start of RL}, communication accuracy is already around 60\% (row 1, column 1) reflecting a relatively high speaking and listening accuracy acquired by the agents at the end of SL
(81\% and 78\% respectively; results not shown in the plots).
When analyzing performance conditioned on meaning ambiguity, we find that communication accuracy before RL is much higher for $M_{test,\ \neg Amb}$ than for $M_{test,\ Amb}$, which was expected and matches the human results of S\&C.
%$\{A, a_{\typeb}, p_{\typea}\}$, are explicit, thus resulting in a much higher starting point (See the green line). 
%For meanings $\{A, a_{\typeb}, p_{\typeb}\}$, two \typeb\ entities have an equal probability of acting as agent or patient and thus are ambiguous if no marker in the utterances, resulting in a lower starting accuracy.
%
Additionally, production preferences before RL (columns 2 and 3, pink cluster around the solid diamond) closely align with the original proportions of the artificial language, reflecting the typical post-SL probability-matching behavior observed in previous work \cite{chaabouni-etal-2019-word,lian-etal-2023-communication}. %Next, we discuss the language drift observed during RL.

\paragraph{Effects of communication}
The increase in overall communication accuracy (orange line) indicates that agents optimize their language during interaction.
This is confirmed by the clear shift in production preferences shown in columns 2 and 3 (first row).
%This is confirmed by the overall production preference shifting from pink to purple, shown in the 2nd subplot.
Specifically, the average preferences after interaction (solid purple circle), indicate a decrease in marker use %\AB{(from 55\% to 35\%)}
alongside a significant shift towards fully using SOV order.
%\AB{(from 66\% to 99\%) Leave numbers in?}
%compared to the initial language.

We further analyze changes in production preferences conditioned on ambiguity. 
For $M_{test,\ \neg Amb}$,
column 2 reveals a general decrease in marker use and a stronger preference for the SOV order.
Additionally, we observe a linear relationship between marker proportion and SOV proportion:
% ($b=0.39,\ SE=0.07,\ t(98)=5.63,\ p<0.001,\ R^2=0.24 $)
the more frequently SOV is used, the more markers are generated
($b=3.62,\ SE=0.31,\ p<0.001$). This could be due to the fact that the input language also has more markers in SOV utterances than for OSV. 
% \lyc{RM this: \sout{However, given that case marking facilitates comprehension, this linear relationship could also suggest that utterances with OSV order have less need for an added marker to be understood: when generating an utterance in the OSV order for $M_{test,\ \neg Amb}$ with a format of $\{A, a_{\typeb}, p_{\typea}\}$, the first token is the unambiguous noun, which is already interpretable.}}
%
For the remaining meanings %$\{A, a_{\typeb}, p_{\typeb}\}$ \lyc{or 
($M_{test,\ Amb}$) we observe an even stronger preference for the SOV order, together with a decrease in marker use, as shown in column 3. 
This suggests that, after communication, agents resolve ambiguity by regularizing the word order to SOV, instead of increasing marker use, which in turn may reflect a general tendency to amplify biases in the original language.

To investigate whether a DOM effect emerges in the productions of neural agents, 
we visualize the individual-level production differences between ambiguous and unambiguous patients (column 4).
On average, we observe 
%On average, agents exhibit 
only a small, but significant, difference in marker usage. While there is a general decrease in marker use, agents retain more markers ($b=0.25,\ SE=0.08,\ p<0.01$) for $M_{test,\ Amb}$.
% (5\% more for ambiguous patients), 
A more noticeable difference in word order preference is observed:
% (19\% more SOV for ambiguous patients).
after communication, agents regularize more towards SOV ($b=3.43,\ SE=0.22,\ p<0.001$) on $M_{test,\ Amb}$ as compared to $M_{test,\ \neg Amb}$. Typicality therefore has significant effects on both case marking and word order.

\subsubsection{Comparison to human results}
While human participants in S\&C tend to increase marker use during communication, we observe a general decrease in marker use in our agent interactions. Instead of producing more markers, the agents tend to regularize towards one consistent  word order to disambiguate the meanings. Even though human learners in S\&C also frequently started over-producing one order versus the other, they still introduced more markers in communication to increase the chance of being understood. Despite these differences, we do see a human-like DOM effect appear during agent interactions, where markers are used significantly more frequently for $M_{test,\ Amb}$, just like the increased marker usage of human participants for animate objects in S\&C.

Another notable difference concerns whether marker use is conditioned on word order. In the language design we adopted from S\&C, the initial case marking proportion is higher for SOV than OSV sentences. Human participants did not maintain these differences while learning the 60\%-SOV with 60\% marking language, but as shown above our agents keep conditioning marker use on word order even during communication. Since the agents seem to be more sensitive to existing patterns present in the initial language, we continue our analyses with the two languages with neutral word order and no conditioning of marker use on word order.

%Although we observe a similar effect of typically on casemarking after communication, 
%the overall production preference is not fully matched with the human results reported in S\&C.
%Specifically, at the beginning of the human learning phase, animate objects (corresponding to $M_{test,\ Amb}$) exhibit a higher use of case marking, but this preference decreases over time, while neural learners show a strong typicality effect on word order.  
%Furthermore, the initial casemarking proportion is higher for SOV than OSV according to the language design. However, S\&C claims no evidence of this uneven marker distribution is found in human learners' production over all-days of training.
%As for the neural learners, significant effects of word order on casemarking appear both before ($b=0.47,\ SE=0.03,\ p<0.001$) and after communication($b=3.05,\ SE=0.30,\ p<0.001$). 
%Based on the above facts, our neural learners are proved to be more sensitive to the word-order than human learners.
%Therefore, it is essential to explore neutral word-order language within this neural framework.

%\section{Learning and Communication \\using Unbiased Language}
\subsection{Neutral Order Languages}

% In the previous experiment, 
% since the initial language is biased towards using more SOV order and assigning a higher marker proportion to SOV compared to OSV, 
% this result is not sufficient to represent the inherent bias of the neural agents.
% In this section, we further investigate
% an Object-marking language $l_{ntl-Obj}$ and a Subject-marking language $l_{ntl-Subj}$, both with neural order
% % , namely 50\% SOV and 50\%OSV,
% and 67\% case marking proportion on each order. 
% \footnote{In preliminary experiments with initially 50\% case marking languages, a strong regularization of not using markers occurred, making it impossible to explore the differential case marking effect.}
% The two languages only differ in the marking constituent, namely Object-marking language ($L_{Obj}[50_{SOV}\&(67_{Smk}+67_{Omk})]$) and Subject-marking language ($L_{Subj}[50_{SOV}\&(67_{Smk}+67_{Omk})]$).

Results for the languages with initial 50/50\% SOV/OSV order are shown in the \textbf{second and third rows} of Figure~\ref{fig:comm_result} (object-marking and subject-marking version, respectively).
%\subsubsection{After SL}
\textbf{Before the start of RL}, communication accuracies for both languages are very similar to those of the dominant order language, and so are the production preferences, again reflecting a probability matching behavior.

%$M_{test,\ \neg Amb}$ result in higher accuracies than $M_{test,\ Amb}$, and the probability matching phenomena are found in the production preferences, similar to the previous experiment.

\subsubsection{Effects of communication}
% The overall production preference shows a different view than the previous one. 
Starting with the object-marking language (second row), 
we again find a drop in the marking proportion, which is present for both ambiguous and unambiguous meanings (columns 2 and 3), but again markers are retained more for $M_{test,\ Amb}$ ($b = 0.49,\ SE = 0.10,\ p<0.001$). 
%(69\% to 49\%),
The development of word order is very different from what was observed for the dominant order object-marking language. Instead of amplifying the already present majority of SOV in the dominant language, agents exposed to the neutral object-marking language develop a clear preference for OSV, which is significantly stronger for $M_{test,\ \neg Amb}$ ($b = 2.75,\ SE = 0.17,\ p<0.001$) as compared to $M_{test,\ Amb}$. In sum, we again see that typicality has a significant effect on both order and case marking.
%Specifically to OSV majority, which is different than $l_{dom-Obj}$\lyc{dominant order language?}.
% (51\%SOV to 27\%SOV). 
%\AB{Does it make sense to discuss the general PP if we only show the breakdown by amb/non-amb?}
%\lyc{amb/non-amb are moving to the same preference, so ok to mention that according to col2\&3? also if we would like to mention the different order preference compared to dom-obj, we need general order preference?}
%
% for meanings $\{A, a_{\typeb}, p_{\typea}\}$,
% despite the order regularization and marker decrease, 
% no linear relationship was found between word order and case marking.
% unambiguous entities, Column 2 shows a strong regularization to OSV,
% % (from 51\% to 16\% SOV order), 
% along with generally decreased use of case marking.
% % (73\% to 45\% case marking).
% Unlike the previous experiment with a linear relationship found, 
% the distribution regularizes more towards OSV\% with a diverse order conditioned marking proportion.
%
A linear relation between word order and marker use appears for $M_{test,\ Amb}$, where less markers are used when SOV is more frequent ($b=-1.98,\ SE=0.15,\ p<0.001$).

%although the averaged preference shows a close distribution around the initial value,
% (52\% to 46\% marking proportion, 54\% to 65\% SOV order proportion), 
%individual samples spread around the whole x-y axes and show a significant negative linear relationship between SOV\% and case marking  ($b=-1.98,\ SE=0.15,\ p<0.001$).
% ($b=-0.51,\ SE=0.06,\ t(98)=-8.82,\ p<0.001,\ R^2=0.44 $).
% A negative linear relationship between SOV\% and case marking indicates that OSV\% order for $\{A, a_{\typeb}, p_{\typeb}\}$ is harder to learn.
% Comparing with the previous experiment where SOV order is marked with the same case marking level of 67\%,
% ambiugous meanings in this unbiased language result in a conditional case markning strategy, instead of only regularizing to the preferred order. 
%
%The difference in production preferences (column 4), 
%together with the logistic mixed effects regression \AB{is this the right naming?}\lyc{or logit regression }, show that typicality has a significant effect on both order and casemarking. 
%Specifically, 
%$M_{test,\ Amb}$ leads to a higher case marking level ($b = 0.49,\ SE = 0.10,\ p<0.001$), 
%while $M_{test,\ \neg Amb}$ regularizes more towards OSV ($b = 2.75,\ SE = 0.17,\ p<0.001$).

% , leading to a weak but human-like differential case marking and un-human-like strong differential ordering. 

%\subsection{Differential Subject Marking}

As expected, results for the subject-marking language (third row) show a symmetric trend where, again, markers persist significantly more for
$M_{test,\ Amb}$ ($b = 0.70,\ SE = 0.10,\ p<0.001$), but the order preference is reversed, where SOV is used for $M_{test,\ \neg Amb}$ significantly more ($b = 3.05,\ SE = 0.25,\ p<0.001$) than for $M_{test,\ Amb}$. These contrasting order preferences between the neutral object-marking and subject-marking languages seem to indicate an agent preference to put the marked entity first.  In addition, the relation between word order and marker use for $M_{test,\ Amb}$ is reversed for the subject-marking language, with more markers used when SOV is more frequent ($b=1.72,\ SE=0.17,\ p<0.001$). Interestingly, FJN similarly found that markers were used more frequently with SOV in their subject-marking language in the early stages of learning, while this was the case for OSV in the object-marking language.
% \lyc{
Since there is no order-conditioned case marking in our neutral languages, these linear relationships could suggest that 
generating an utterance for $M_{test,\ Amb}$ in the majority order creates a need for an added marker to be reliably understood, while using the other order serves, in itself, as a way to disambiguate.

\section{Discussion and Conclusion}

We used \frameworkx\ to study the emergence of Differential Case Marking, 
employing previous experimental setups of human studies by FJN and S\&C.
Neural agents do not have the same biases in learning and signal production as humans, so different preferences between agents and humans after learning and communication are expected. Indeed, we saw that our agents were more sensitive to specific patterns in the input language than humans, and had a greater tendency to drop markers and disambiguate meanings using word order. While our agents learned about typicality of entities solely based on statistical properties in the artificial language, human participants in FJN and S\&C already had knowledge about animacy in addition to this. Moreover, human participants have existing preferences to iconically relate marked forms with atypical events
\cite{aissen2003differential}, 
while agents have no such bias. Finally, humans have a preference to place human and animate entities before inanimates in a sentence \cite{aissen2003differential}, while our agents are not aware of these distinctions. The interacting effects of all these biases can make it difficult to tease apart causal mechanisms contributing to the emergence of DCM in human experiments.
% when working with human participants
As discussed in the introduction, FJN and S\&C indeed found conflicting results when looking at the role of learning. Complementing these previous findings, and supporting S\&C's conclusions, our simulations demonstrate that DCM does not arise as the result of learning, but does emerge when agents start communicating in pairs. Importantly, our agent setup allowed to study these factors in the absence of prior language experience and sense of animacy or iconicity in the learners.

Beyond replicating human ALL results with linguistically naïve neural learners, employing NeLLCom also offers advantages in scalability. Using neural agents, we can conduct numerous iterations and explore diverse language conditions, which would be costly and time-consuming with human participants. For example, studying real communication between two or more interacting participants would have been hard to coordinate with the large number of (online) participants included in S\&C's study, which may explain their use of a chatbot. In our setup, we could easily model pairs of interacting agents, and this can just as easily be extended to groups. 
Additional directions for future work include experimenting with a less clear-cut distinction between entity-role distributions (e.g. 55/45\% and 5\%/95\%, rather than 50/50\% and 0/100\%), %e.g. `entity-5' occurring as patient 90\% of the time, 
which would more closely resemble real language distributions. Another way to possibly achieve more human-like patterns would be to endow agents with a notion of animacy by initializing them with meaning embeddings pre-trained on large text corpora. 

To conclude, NeLLCom-X can be used to complement experimental research on language evolution, allowing us to precisely control and compare various aspects of language systems and population dynamics while at the same time revealing ways in which neural agent learning and language use differs from that of humans.

\bibliographystyle{apacite}

\setlength{\bibleftmargin}{.125in}
\setlength{\bibindent}{-\bibleftmargin}

\bibliography{main}

\end{document}

%% file: contents/lang_table_v4.tex
\begin{center}
\caption{\label{tab:language-table} The miniature languages used in this study.}
% and the corresponding utterances for meaning $m$=\{$A$: \textsc{make}, $a$: \textsc{baker}, $p$: \textsc{cake}\}}
\label{tab:lang}
% \vskip 0.12in
\begin{tabular}{@{}c@{\ } | c@{\ } c@{\ \ } c@{\ \ } c@{}}
%\hline
language & 
mark & \textsc{sov} & $mk|$\textsc{sov} & $mk|$\textsc{osv} \\
\hline
% dominant-order (S\&C, exp.1) & obj & 60\% & 67\% & 50\% \\
dominant-order & \multirow{2}*{obj} & \multirow{2}*{60\%} & \multirow{2}*{67\%} & \multirow{2}*{50\%} \\
(S\&C, exp.1) & & & &  \\
\hline
\multirow{2}{*}{neutral-order} & obj & 50\% & 67\% & 67\% \\
 & subj & 50\% & 67\% & 67\% \\ 
\hline
\end{tabular} 
\end{center}

% OTHER EXAMPLE LANGUAGES NOT USED IN THE EXPS:
% O & 100\% & 50\% & --- \\

%\hline
% O & 50\% & 0\% & 100\% 
%  & \it Baker Cake Make | Cake mk Baker Make \\
% %\hline
% S & 0\% & --- & 50\%
%  & \it Cake Baker Make |  Cake Baker mk Make \\
% %\hline
% S & 50\% & 100\% & 0\%
%  & \it baker mk cake make | cake baker make \\
 

%% file: contents/fig_merge_v2.tex
  \centering
  \begin{tabular}{c@{\hspace{.8mm}}  c  c  c  c }

    \multirow{5}{*}{\makebox[0pt][l]{\hspace{0.20cm}\rule{0.2pt}{58ex}}} &  
    \multirow{2}{*}{\small \sc{Communication success}} &
   \multicolumn{3}{c}{\sc{Production preferences}}\\

   % \cline{3-5}
   &
   \multicolumn{4}{l}{\makebox[10cm][l]{\hspace{4.6cm}\rule{12.6cm}{0.2pt}}}
    \\

   & & \small $M_{test,\ \neg Amb}$ 
   & \small $M_{test,\ Amb}$
   & \small $\Delta(M_{test,\ \neg Amb},M_{test,\ Amb})$ \\
   %& \small difference \\
  % \hline

   % % & \small accuracy 
   % & \small \sc{Communication success} 
   % % & \small general 
   % & \small unambiguous 
   % & \small ambiguous
   % & \small difference
   % \\
   % \hline
    
    %\rotatebox[origin=c]{90}{\small $l_{dom-OBJ}$} &
    \rotatebox[origin=c]{90}{\small dominant-order} &
    % \rotatebox[origin=c]{90}{\small $l_{obj}$+60s+}
    % \rotatebox[origin=c]{90}{\small 67smk + 50omk} &

    % \begin{minipage}{\subfigsize\textwidth}
    %   \includegraphics[width=\columnwidth]{fig/v2_with_xylabel/flex-dom-inclusive_osv40_mkOSV50_mkSOV67_typicality_condi_accuracy_ep60.png}
    % \end{minipage}

    \makecell[tc]{\vspace{.5mm}
    \begin{minipage}{\subfigsize\textwidth}
        \includegraphics[trim={0 1.1cm 0 0},clip,width=\columnwidth]{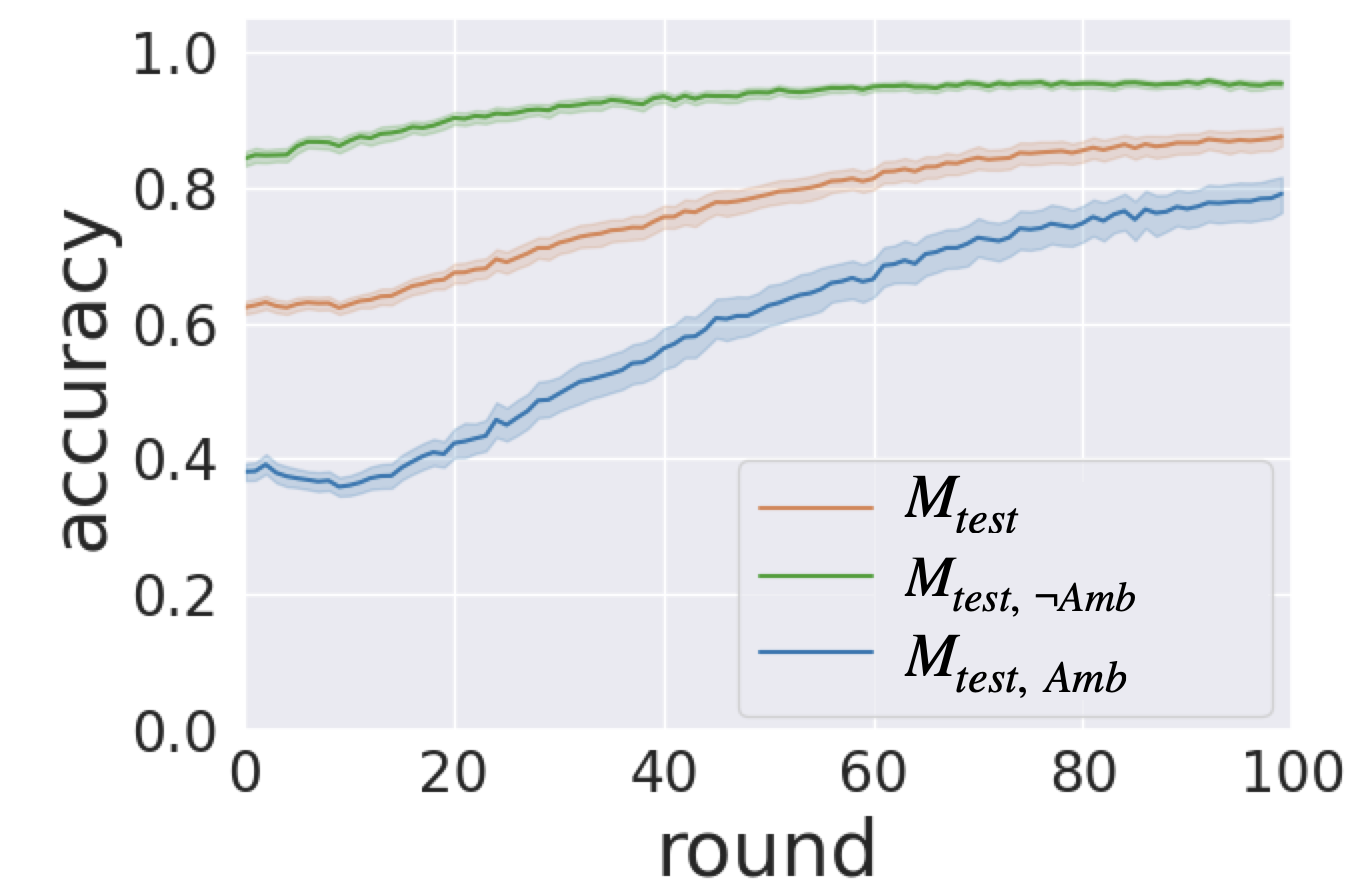}
    \end{minipage}
    \vspace{.5mm}
   }
    &
    \makecell[tc]{\vspace{.5mm}
    \begin{minipage}{\subfigsize\textwidth}
      \includegraphics[trim={0 1.1cm 0 0},clip,width=\columnwidth]{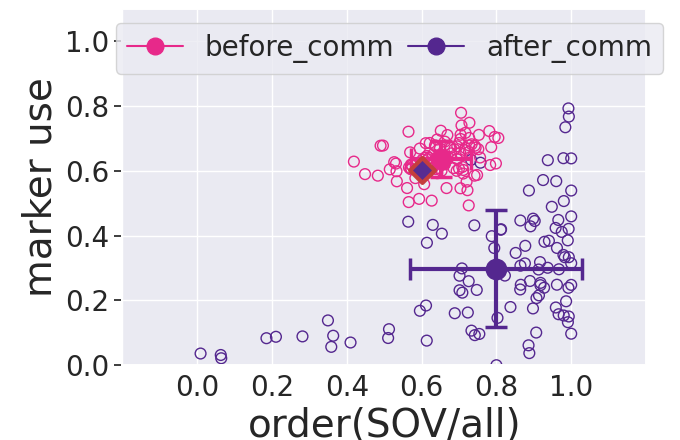}
    \end{minipage}
    \vspace{.5mm}
   }
    &
    \makecell[tc]{\vspace{.5mm}
    \begin{minipage}{\subfigsize\textwidth}
      \includegraphics[trim={0 1.1cm 0 0},clip,width=\columnwidth]{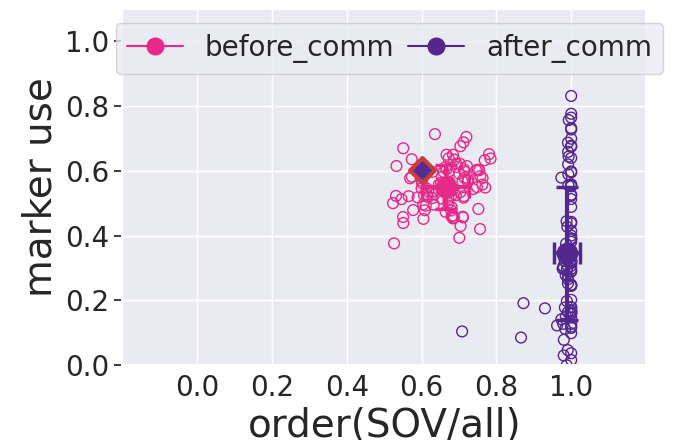}
    \end{minipage}
    \vspace{.5mm}
   }
    &
    \makecell[tc]{\vspace{.5mm}
    \begin{minipage}{\subfigsize\textwidth}
      \includegraphics[trim={0 1.1cm 0 0},clip,width=\columnwidth]{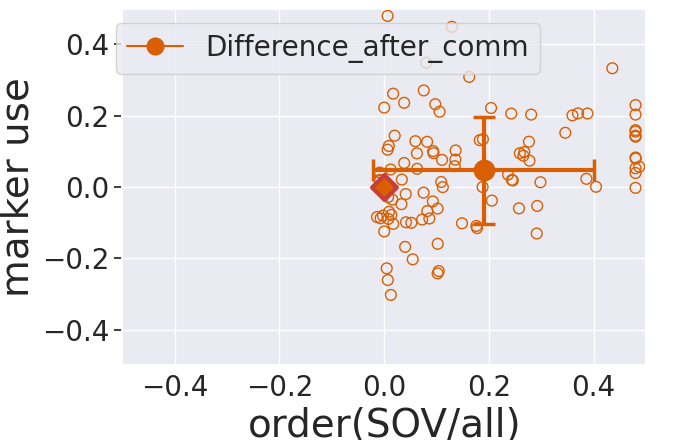}
    \end{minipage}
    \vspace{.5mm}
   }
    \\
   % \hline
    
    %\rotatebox[origin=c]{90}{\small $l_{ntl-OBJ}$} &
    \rotatebox[origin=c]{90}{\small neutral (\textsc{obj})} &
    % \rotatebox[origin=c]{90}{\small $l_{obj}$+50s+}
    % \rotatebox[origin=c]{90}{\small 67smk + 67omk} &
    \makecell[tc]{\vspace{.5mm}
    \begin{minipage}{\subfigsize\textwidth}
      \includegraphics[trim={0 1.1cm 0 0},clip,width=\columnwidth]{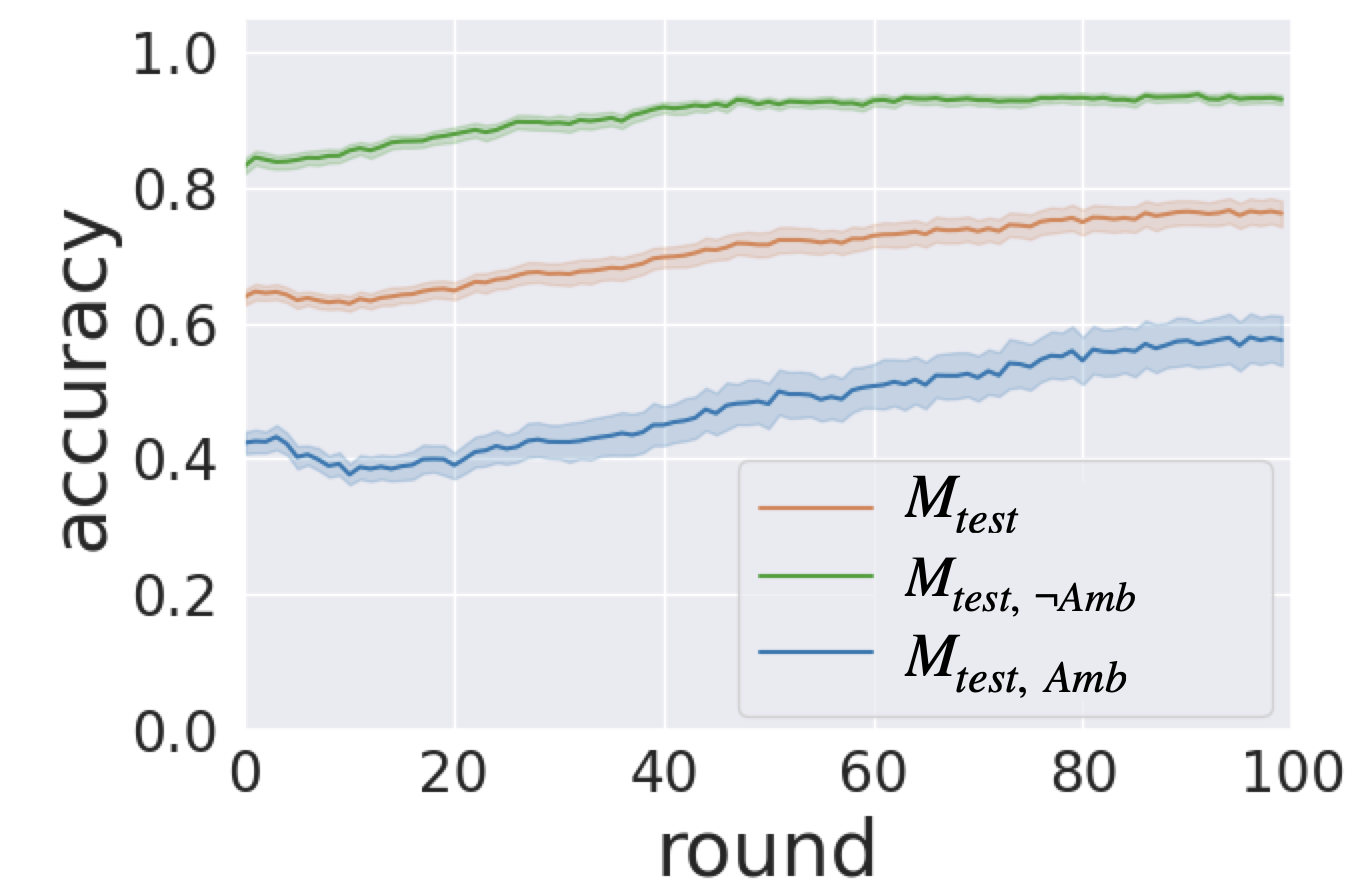}
    \end{minipage}
    \vspace{.5mm}
   }
    &
    \makecell[tc]{\vspace{.5mm}
    \begin{minipage}{\subfigsize\textwidth}
      \includegraphics[trim={0 1.1cm 0 0},clip,width=\columnwidth]{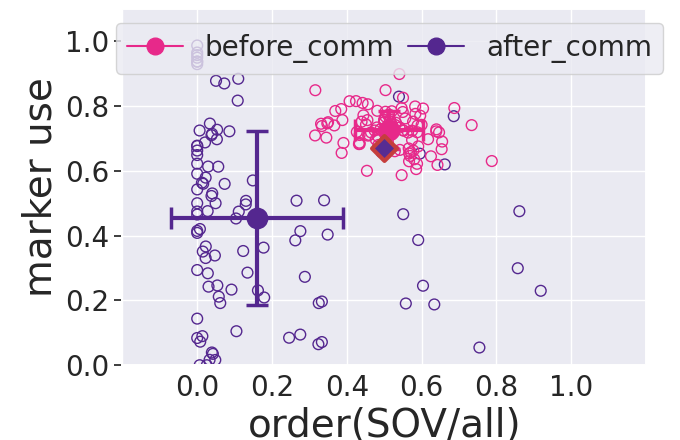}
    \end{minipage}
    \vspace{.5mm}
   }
    &
    \makecell[tc]{\vspace{.5mm}
    \begin{minipage}{\subfigsize\textwidth}
      \includegraphics[trim={0 1.1cm 0 0},clip,width=\columnwidth]{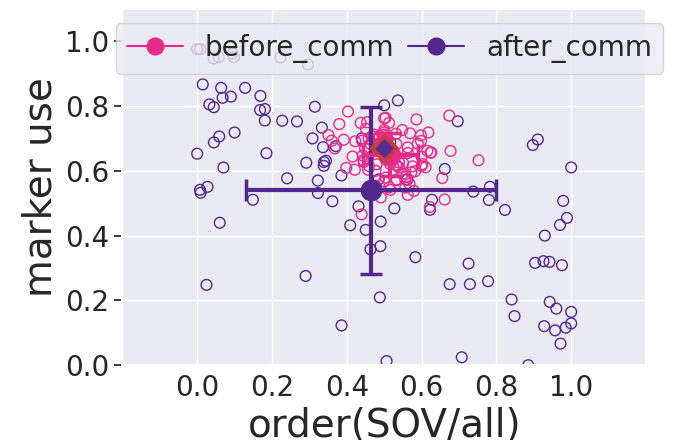}
    \end{minipage}
    \vspace{.5mm}
   }
    &
    \makecell[tc]{\vspace{.5mm}
    \begin{minipage}{\subfigsize\textwidth}
      \includegraphics[trim={0 1.1cm 0 0},clip,width=\columnwidth]{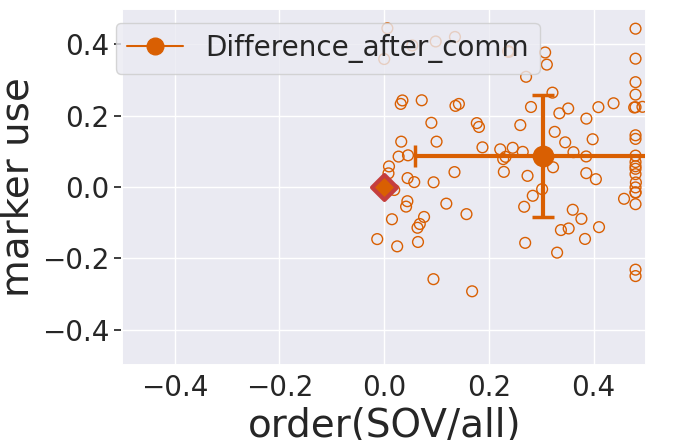}
    \end{minipage}
    \vspace{.5mm}
   }
    \\
   % \hline
    
    %\rotatebox[origin=c]{90}{\small $l_{ntl-SUBJ}$} &
    \rotatebox[origin=c]{90}{\small neutral (\textsc{subj})} &
    % \rotatebox[origin=c]{90}{\small $l_{subj}$+50s+}
    % \rotatebox[origin=c]{90}{\small 67smk + 67omk} &
    \makecell[tc]{\vspace{.5mm}
    \begin{minipage}{\subfigsize\textwidth}
      \includegraphics[width=\columnwidth]{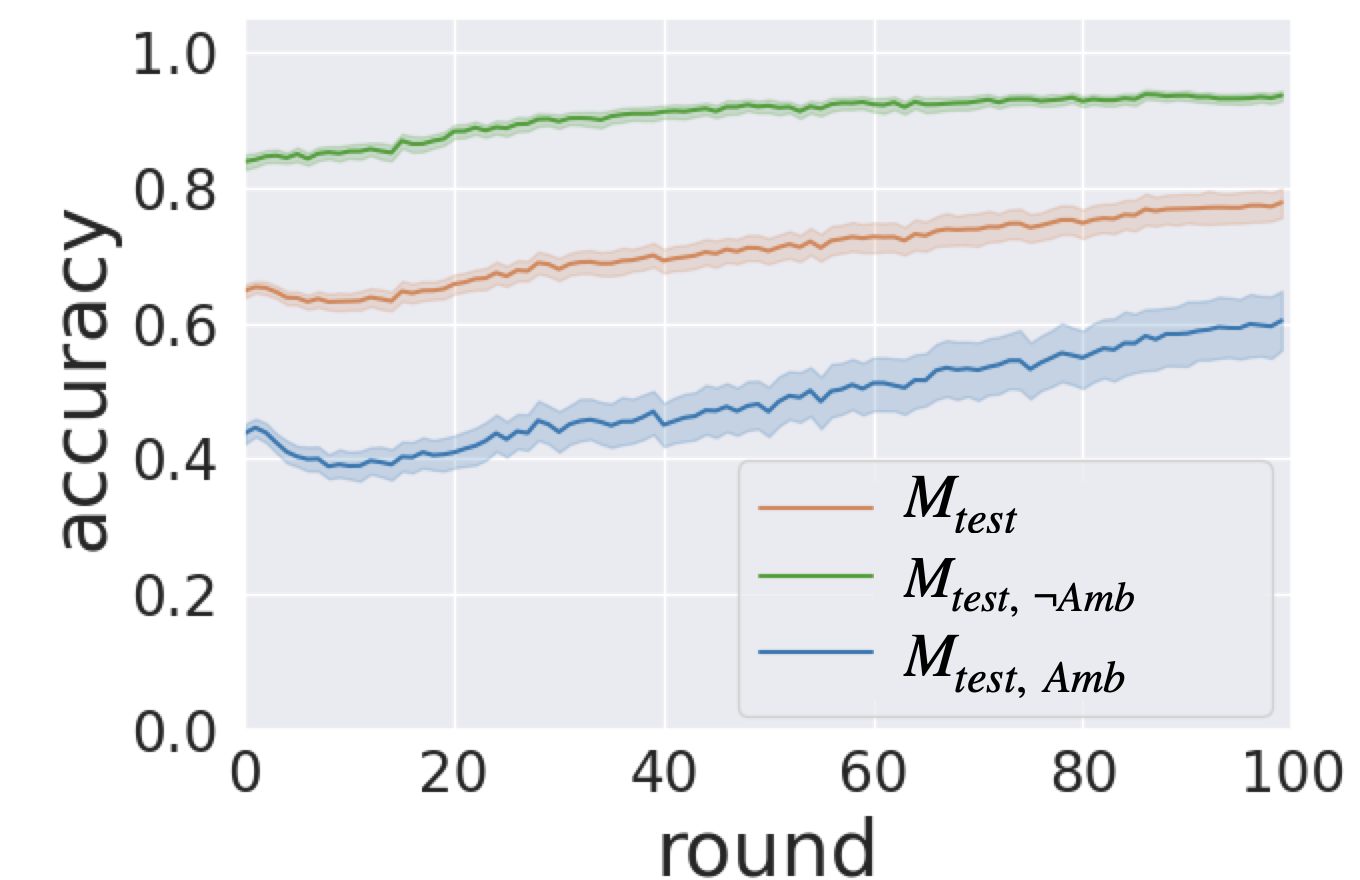}
    \end{minipage}
    \vspace{.5mm}
   }
    &
    \makecell[tc]{\vspace{.5mm}
    \begin{minipage}{\subfigsize\textwidth}
      \includegraphics[width=\columnwidth]{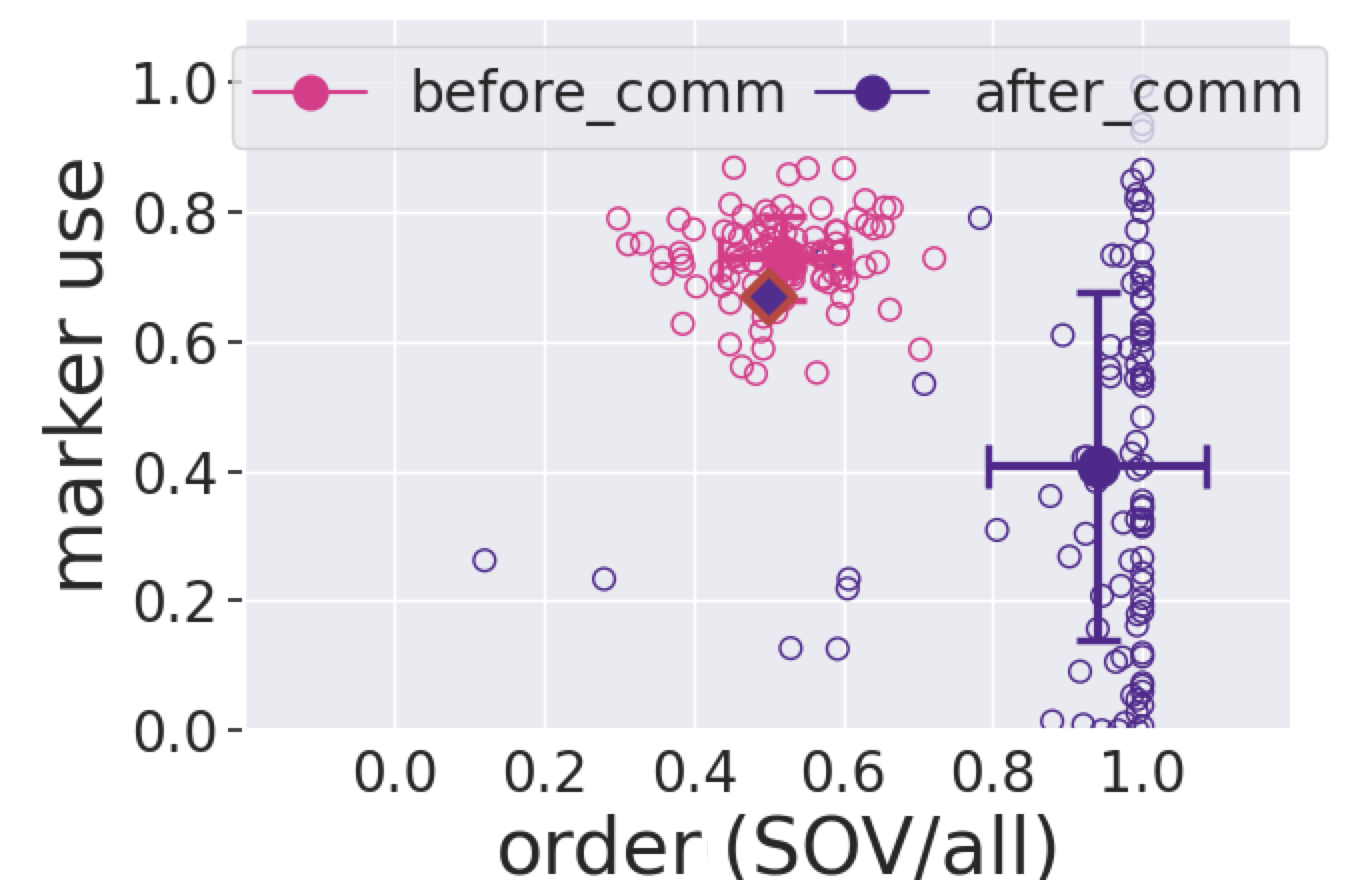}
    \end{minipage}
    \vspace{.5mm}
   }
    &
    \makecell[tc]{\vspace{.5mm}
    \begin{minipage}{\subfigsize\textwidth}
      \includegraphics[width=\columnwidth]{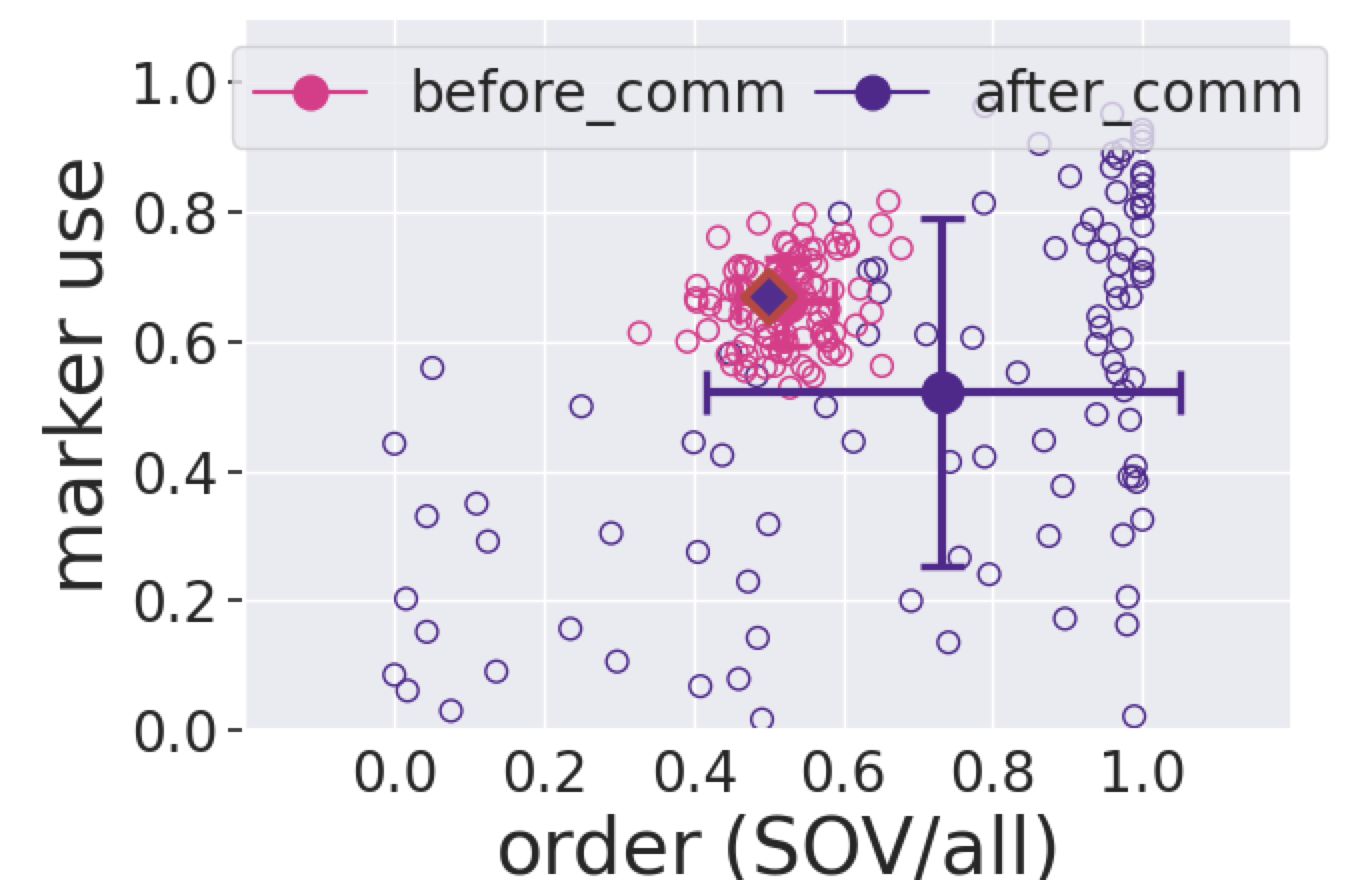}
    \end{minipage}
    \vspace{.5mm}
   }
    &
    \makecell[tc]{\vspace{.5mm}
    \begin{minipage}{\subfigsize\textwidth}
      \includegraphics[width=\columnwidth]{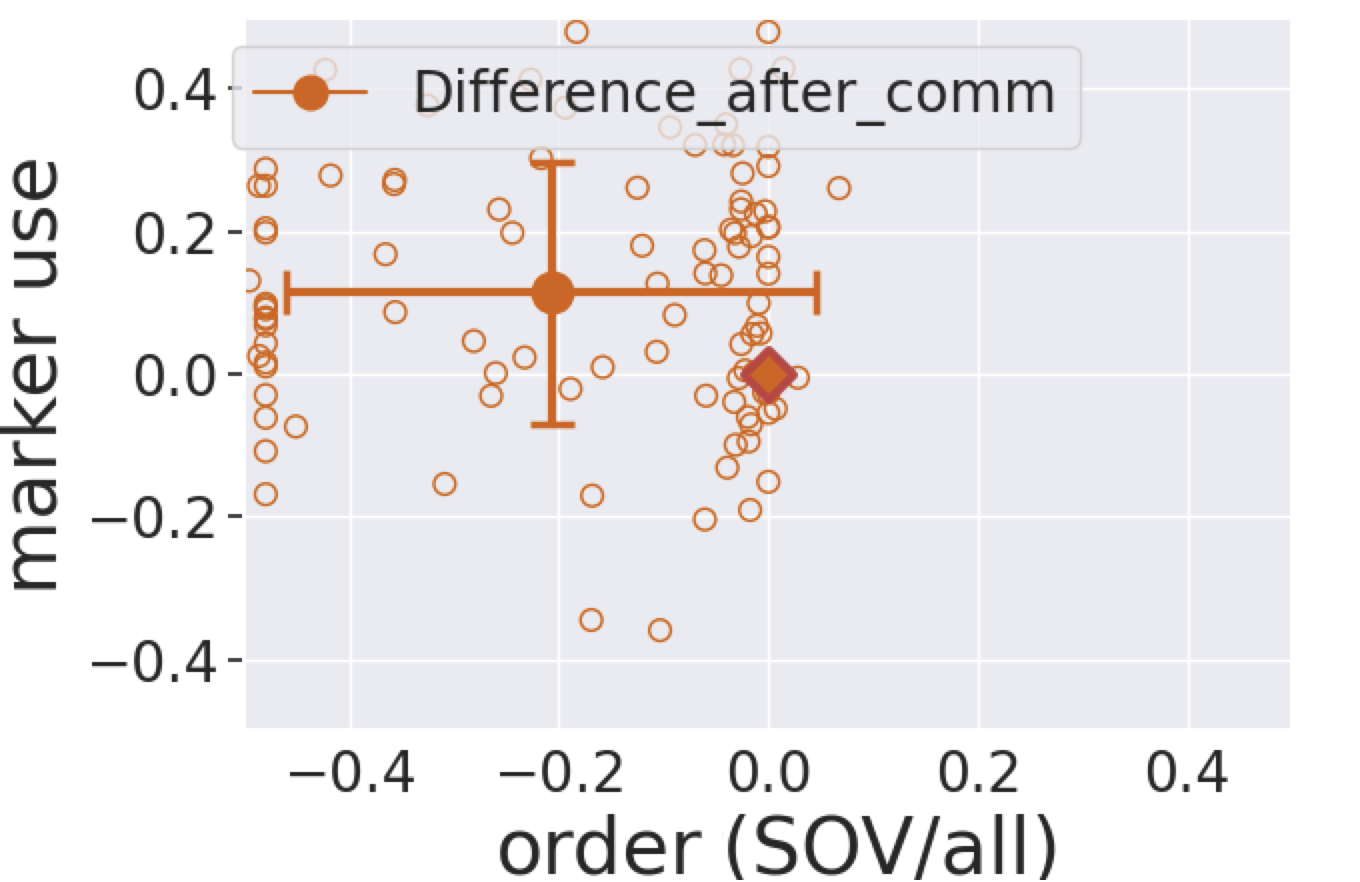}
    \end{minipage}
    \vspace{.5mm}
   }
    \\
    %\hline
    
  \end{tabular}